# LULC Segmentation of RGB Satellite Image Using FCN-8

Abu Bakar Siddik Nayem, Anis Sarker, Ovi Paul, Amin Ali, Md. Ashraful Amin, and AKM Mahbubur Rahman

Independent University Bangladesh
{1510190, 1521745, 1531144, aminali, aminmdashraful, akmmrahman}
@iub.edu.bd

**Abstract**- This work presents use of Fully Convolutional Network (FCN-8) for semantic segmentation of high-resolution RGB earth surface satel-lite images into land use land cover (LULC) categories. Speci cally, we propose a non-overlapping grid-based approach to train a Fully Convo-lutional Network (FCN-8) with vgg-16 weights to segment satellite im-ages into four (forest, built-up, farmland and water) classes. The FCN-8 semantically projects the discriminating features in lower resolution learned by the encoder onto the pixel space in higher resolution to get a dense classi cation. We experimented the proposed system with Gaofen-2 image dataset, that contains 150 images of over 60 di erent cities in china. For comparison, we used available ground-truth along with images segmented using a widely used commeriial GIS software called eCogni-tion. With the proposed non-overlapping grid-based approach, FCN-8 obtains signi cantly improved performance, than the eCognition soft-ware. Our model achieves average accuracy of 91.0% and average Inter-section over Union (IoU) of 0.84. In contrast, eCognitions average accu-racy is 74.0% and IoU is 0.60. This paper also reports a detail analysis of errors occurred at the LULC boundary.

## 1 Introduction

Efficient land management tasks such as change detection, urban planning, re-source monitoring, environmental protection, agriculture, building road maps, planning for socioeconomic development etc. [4], [3], [2], [6] depends on proper identi cation of LULC. Usually classi cation of LULC is performed manually on map images using geographic information system (GIS) softwares (e.g. eCog-nition [13]), which is a time-consuming and ine cient approach. Some Re-searchers propose machine learning techniques to perform automatic classi ca-tion of LULC in di erent semantic classes (i.e forest, farmland, water, built-up area, meadows) from satellite images. Satellite images have a number of channels i.e. R, G, B, Near Infra-red (NIR), and Infra-red (IR) etc. From these channels, geographic researchers attempt to calculate normalized di erence vegetation index (NDVI)and normalized di erence moisture index (NDMI) for forest and farmland segmentation. They also derived re ectance index (RI), brightness in-dex (BI) and some other indexes to identify other classes. These indexes are calculated using speci c formulas derived for each class/index over the various combination of raster channels. The bands representing NIR(Near Infrared) and RED will vary from satellite to satellite. Often, a satellite may give only to the ones required for such index calculations rather than giving the exact spec-trum. In that case, the band closest to the required one is used to perform such calculations. To automatically identify vegetation in satellite images, Cheng et al. [4] used Histogram of Oriented Gradients (HOG), Scale-Invariant Feature Transform (SIFT), and Local Binary Pattern (LBP) for feature extraction and support vector machine as classi er. While tested on remote sensing image data provided by Multi-resolution land cover characteristics (MRLC) consortium [9], their system showed 79.6% overall accuracy.

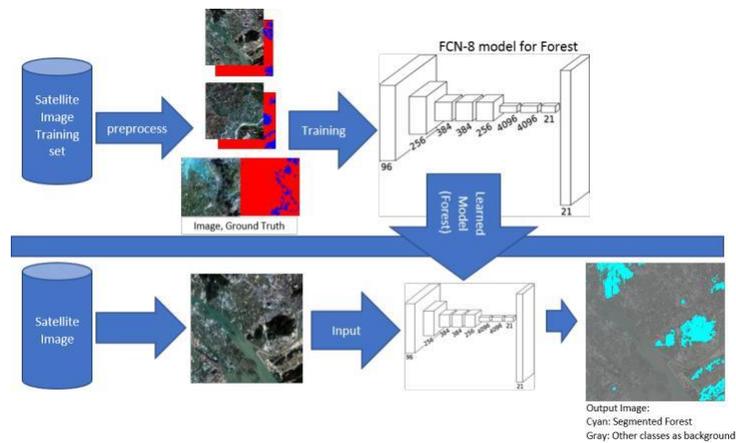

Fig. 1. Block Diagram of the proposed approach (Forest model). Upper part is the training process. Lower part is the prediction stage.

Recent researches use various deep learning methods with better success [5], [10], [7], [12] for land cover classi cation and segmentation. Ben Hamida et al. [5] uses DenseNet [14] and SegNet[1] for ne segmentation and coarse segmentation, respectively on multi-spectral Sentinel-2 images. They report overall accuracy of 51.4% with DenseNet and of 83.9% with SegNet on GlobeCover data. Pira-manayagam et al. [10] introduced early and late fusion of features in a neural architecture(Fully Convolutional Network [8], FCN) for application in multisen-sor aerial/satellite image classi cation. They achieved overall accuracy. 59.87% and average F1 score. 0.51 on RGB images from satellites. Shengjie Liu et al. [7] uses Object-based image analysis (OBIA) for land use and land cover mapping using optical and synthetic aperture radar (SAR) images. To obtain object-based thematic maps, they developed a new method that integrates object-based post-classification refinement (OBPR) and CNNs for LULC mapping using Sentinel optical and SAR data. They achieved accuracy of 77.64% for the Zhuhai-Macau LCZ dataset with 100 m spatial resolution. Above performances are not sat-isfactory for the automatic LULC classification due to some limitations. The limitations include. scarcity of ground truth data, ambiguity of boundary pixels, inability to model spatial characteristics of LULC classes etc.

To improve the segmentation map, Tong et al. [12] used an ensemble of patch-wise classifier with hierarchical segmentation method. Later, they used selective search to estimate the boundary. However, patch based systems suffer a number of drawbacks for satellite images. Firstly, patch-based technique will limit the segmentation process on objects that might have a wide range of shape, size, and densities. Secondly, the error in detecting the patches will propagate to the next level where patches are united together to create final segmentation. Thirdly, patch based systems achieves very poor generalizability. To overcome the above mentioned limitations, we employ a deep learning-based semantic segmentation method that segments forest, built-up, farmland, and water area from the satellite RGB images directly. As RGB images are easily explainable and observable for LULC classes, we focus our segmentation on RGB image only.

Particularly, we employ a pre-trained convolutional neural network with VGG-16 weights as encoder to get the Fully Convolutional Network (FCN-8)[8]. Figure 1 shows the block diagram for forest segmentation where upper part de-picts the training process and lower part represents prediction stage. During training, we supply RGB images with corresponding ground truth to learn the FCN-8 model for forest. In the prediction part, we segment the input image for forest area using the learned model. We do the same approach for other LULC classes. In initial experiments, we downsampled the satellite image of size 7200 6800 to 224x224. Huge downsampling process induce unexpected errors in the boundary. To improve the performance, we divide the full size satellite image into 224x224 non-overlapping sub-images. Hence, the resolution of the input image is preserved while fed to the FCN-8. This approach achieves great success with an average accuracy of 91% and an average Intersection over Union (IoU) of 0.84 while segmenting the RGB images from Gaofen-2 Image Dataset (GID) dataset [12]. Moreover, the FCN-8 model outperforms eCognition [13]. Detail performance comparison is done in the chapter 4. Hence, the proposed automated LULC process with FCN-8 described in this paper can help to ex-tract vital information to understand our planet better. However, sometimes, the FCN-8 induces errors in finding the boundaries, especially for farmlands which encompass great diversity in color and shape. We did a detail error analysis where we described the possible sources of these errors.

## 2  Dataset and Methodology

This section describes the dataset and the preprocessing part. Also, we describe the approach and methodology we use in this paper.

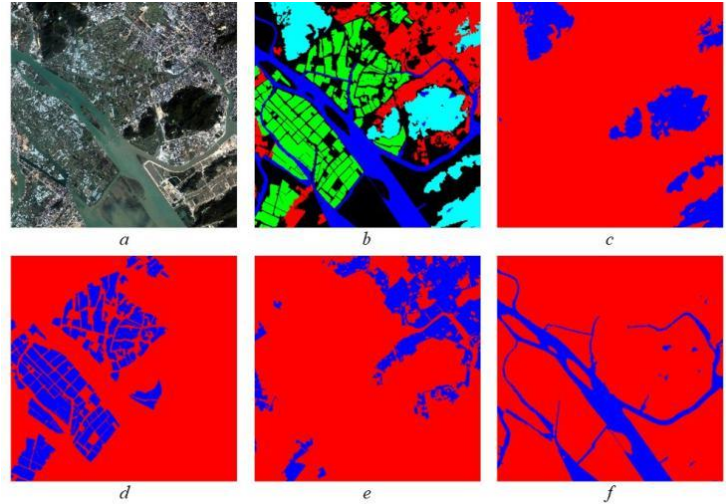

Fig. 2. Sample input image [a]. [b] red pixels represent Build-up class; cyan ones repre-sent forest, green ones represent farmland and nally, blue ones represent water. black pixels are unrecognized by the authors. Images [c], [d], [e] and [f] show the binary ground truth image for Forest, Farmland, Built-up, and Water class

### 2.1  Dataset Description

The GID dataset contains a total of 150 images in tif format of resolution 7168x 6720. Each pixel covers upto 4 meters resolution.GID provide a spectral range of blue (0.45-0.52 µm), green (0.52-0.59 µm), red (0.63-0.69 µm) and near-infrared (0.77-0.89 µm), and a spatial dimension of 7168x6720 pixels covering a geographic area of 506 km$^2$.

Ground Truth The dataset also contains ground truth labels for these 150 images. The ground truth images contain di erent colored pixels to show four di erent LULC classes. Figure 2a and 2b is an example of input image and corresponding ground truth that are provided by GID dataset. However, black pixels represent unrecognized area by the authors of [12]. Our experiments do not include any unrecognized area.

Preprocessing We preprocess the ground truth images to produce binary image for each class. For each class, we make the pixels in target class into blue and the rest of the classes into red. Hence, we produce four binary ground truth images for each input image. Binary images for Forest, Farmland, Built-up, and Water classes are shown in 2c - 2f.

Data Augmentation We used nine (9) augmentation methods for our data augmentation. Hence, each of the images were increased to 10 different images.

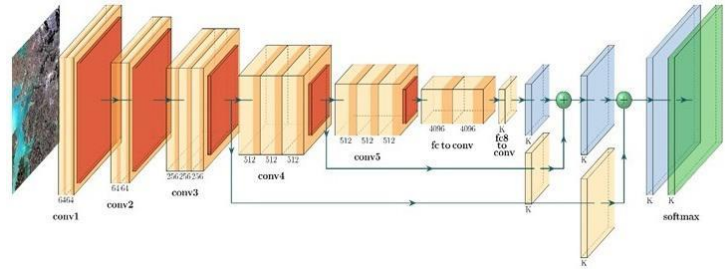

Fig. 3. Fully Convolutional Network FCN - 8

including the original image. We ipped the images vertically, horizontally, rotate the images into anti-clockwise 90 , 180 and 270 . Then we used the contrast stretching, gamma, and hue.

## 2.2 LULC with Semantic Segmentation

LULC in a very high level a classi cation task. It necessitates the most accurate classi cation. Semantic segmentation is required which classi es the image on the pixel level, that is, each pixel in the image belongs to a class. In remote sensing paradigm, semantic segmentation indicates segment/detection of area that are consists of same kind of land cover. For example, semantic segmentation for forest will give us all forest area regardless of any size, shape, and texture.

## 2.3 VGG-16 architecture and Fully Convolutional Layer.

The VGG network architecture, a convolutional neural network (CNN), was introduced by Simonyan and Zisserman et al. [11]. Using a convolutional layer as the last layer of the VGG-16, the resulting fully convolutional network (FCN-8 [8]) segments the image instead of classifying it. Pixels belonging to same class given the same color label. Generally, CNN is connected networks of convolution layer and pool layer where convolution layers are used to encode the lower level of features to the higher level of semantic abstract. Pooling layers are used to decrease the dimension of the higher levels. Reducing volume size is handled by max pooling. For a CNN, the input is an image X of dimension m n d(e.g., 3 color channels(R,G,B)) . The neurons are also arranged in 3 dimensions. Each neuron is connected to a number of inputs in the previous layer.

The features obtained are further passed through multiple convolution, ReLU, and subsampling or pooling layers. These layers are then followed by a fully connected convolutional layer. This transformation allows the network to generate coarse maps with spatial support. A convolution transpose layer [8] is then used to bring the coarse map to the original image resolution.

## 3 Experimental Setup and Results

In this section, we discuss our experimental setup, train/test splits, training-testing process, and the evaluation metrics.

### 3.1 Train and Test Splits

The dataset contains 150 images in total. All of them are RGB images. However, each of the LULC classes are not present in every image. In order to make good representation of LULC classes in train and test sets for our experiments, we considered the images with at least 5% pixels belonging to one class for training and testing the binary model for that particular class. We have trained our binary FCN-8 models separately for each of the class. The table 1 shows that in 31 images, Forest was present in at least 5% of the area. Similarly, Farmland, Built up and Water is present in 131, 60 and 72 images respectively. We did split our train data and test data.

Table 1. Number of images containing at least 5% of each class

| Class | Forest | | Farmland | | Builtup | | Water | |
|---|---|---|---|---|---|---|---|---|
| Total images | 31 | | 131 | | 60 | | 72 | |
| Train and Test Split | 25 | 6 | 119 | 12 | 52 | 8 | 63 | 9 |

About 8%-20% images from each class are used as test set depending on the dataset size for each class as of Table 1. Numbers in shaded red cells represent the number of images kept separate for nal testing and evaluation for each class and Numbers in shaded green represent the training set size. In the nal test process, we use our test split that consist of all unseen images that we kept separate. Hence, total number of test images separated from all classes, 6+12+8+9=35.

### 3.2 Training and Testing

We train a separate binary model for each of the four LULC classes i.e. forest, water, farmland, built-up. In this training phase, each image is augmented to increase the dataset size. Dataset was augmented in 9 ways as discussed in the augmentation section. Then we test the models on the test splits.

### 3.3 Evaluation Matrix

Calculating confusion matrix gives us the performance of our classi cation model whether it is getting right and the types of error it is making. In our segmen-tation model, the output is a binary image which contains targeted class or non-targeted class. We compare this output image with binary ground-truth im-age. In this case, we use the pixel by pixel Accuracy, Recall,

Precision, F1-score, and Intersection over Union (IoU). Here, IoU is calculated by dividing the area of overlap by the area of union.

We measure these performances for our 4 classes Forest, Built up, Farmland and Water individually.

## 4  Result

In this section, we discuss the performance of our models and show the comparison with eCognition.

### 4.1  Training Process

The training process of FCN-8 involves 100 epochs with learning rate of 0.01. During the train, we downsample the original images from 7168x6720x3 into 224x 224x3 dimensions. We also perform augmentation on the downsampled images to produce nine augmented versions of the original downsample image. Then we supply the downsample images and their augmentation images to the pre-trained FCN-8. We also supply the corresponding binary ground truth image of 224x224x3 for the particular binary segmentation FCN-8 model. In the non-grid-wise manner where we downsample the images, augmentation is used to increase dataset size. But in the grid wise training no augmentation is used as the dataset becomes very large even without augmentation. So, 25x10 = 250 input images along with their ground truth are passed through the network for downsampled training. Therefore, in grid-wise manner, we feed 960x25 = 24000 input images along with 24000 corresponding binary ground truth images for the forest FCN-8 model.

### 4.2  Testing Process

After the training, we supply the test set to the ne-tuned individual FCN - 8 model for each binary class. Similar to the training process, the test dataset passes through the network in a grid-wise manner where each original resolution image is split into 960 images with 224x224 resolution. However, during testing, we did not perform any augmentation on the test set. Hence, 6 input images are tested with Forest FCN-8 model to calculate the performance of this model. We feed 12,8 and 9 test images to Farmland model, to Built-up model, and to Water model respectively as table. 1 . Then, we compare the segmented output with the corresponding binary ground truth data to calculate confusion matrices for each of the model. The performance metrics calculated from the confusion matrices are shown in the table. 2

Table 2. Performance of FCN-8 and eCognition(eCog) on the test set.

| Class | Accuracy | | IoU | | Recall | | Precision | | F-1 | |
|---|---|---|---|---|---|---|---|---|---|---|
| | FCN-8 | eCog | FCN-8 | eCog | FCN-8 | eCog | FCN-8 | eCog | FCN-8 | eCog |
| Forest | 0.82 | 0.80 | 0.73 | 0.77 | 0.30 | 0.63 | 0.81 | 0.72 | 0.40 | 0.65 |
| Builtup | 0.83 | 0.73 | 0.71 | 0.58 | 0.52 | 0.30 | 0.51 | 0.19 | 0.45 | 0.21 |
| Farmland | 0.73 | 0.63 | 0.60 | 0.47 | 0.33 | 0.23 | 0.59 | 0.32 | 0.3 | 0.32 |
| Water | 0.93 | 0.73 | 0.86 | 0.59 | 0.76 | 0.69 | 0.78 | 0.40 | 0.76 | 0.48 |
| Average | 0.85 | 0.74 | 0.76 | 0.60 | 0.43 | 0.46 | 0.75 | 0.41 | 0.48 | 0.42 |

For the test set, our model achieves good accuracy across the classes as we can see in Table 2. The IoU is lower for forest with FCN-8 than the IoU with eCognition. However, the recall scores for FCN-8 are very poor for all classes except Water. That means the sensitivity of the FCN - 8 model is low. Even though the precision score is good for Forest and Water class, the low recall resulted in lower F1 score. The performance metrics for eCognition are showed in the right side in Table 2. Though, the FCN-8 models outperform eCognition, accuracy, IoU, and F1 scores are not high enough.

### 4.3 Grid-wise training and testing

In this section, we perform experiments with the satellite images while keeping the spatial resolution unchanged. However, the size of the FCN-8 is 224 x 224 whereas the input image dimension is 7168 x 6720. Therefore, to keep the image texture unchanged, we divided each full-sized input image into non overlapping sub-images. Each of the sub-images are of size 224 x 224. The sub-images are created by dividing full image grid-wise so that we can stitch them together after segmenting all sub-images. After segmenting all these sub-images, we con-catenate the outputs to produce the nal segmented output of full dimension 7168 x 6720.

Table 3. Performance of FCN-8 on cropped sub-images of test data.

| Class | Accuracy | IoU | Recall | Precision | F-1 |
|---|---|---|---|---|---|
| Forest | 0.915 | 0.847 | 0.565 | 0.901 | 0.640 |
| Builtup | 0.914 | 0.846 | 0.506 | 0.850 | 0.626 |
| Farmland | 0.845 | 0.735 | 0.711 | 0.699 | 0.691 |
| Water | 0.964 | 0.932 | 0.862 | 0.905 | 0.877 |
| Average | 0.910 | 0.840 | 0.661 | 0.839 | 0.708 |

Table 3 shows the detail performance when we test images with the FCN - 8 model that was trained in grid-wise fashion. Water region is segmented more accurately than other classes. Our observation is that water is present in more dense fashion with little or no small parts scattered around in a image where water is present at all. Other classes has scattered small parts all over the image if present i.e. they are not as dense as water is. This leads to slightly less satisfying result for other classes. However, farmland su ers the most from this spatial distribution characteristics resulting in worst performance when compared to other classes.

After training is done, we also create sub-images from the test images and then we pass them to the ne-tuned FCN-8. Then, we stitch the segmented output.

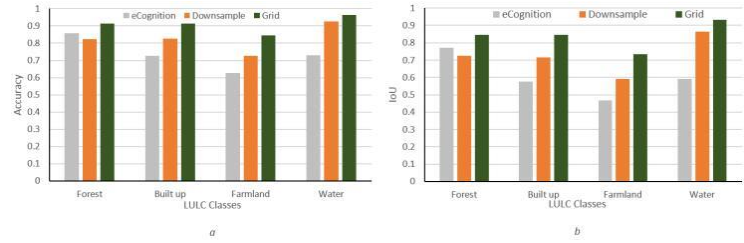

Fig. 4. Performance Comparison for test-image-segmentation by eCognition, FCN-8 with down-sampled images, and FCN-8 with 224 x 224 grid sub-images. (a) Accuracy, (b) Intersection over Union.

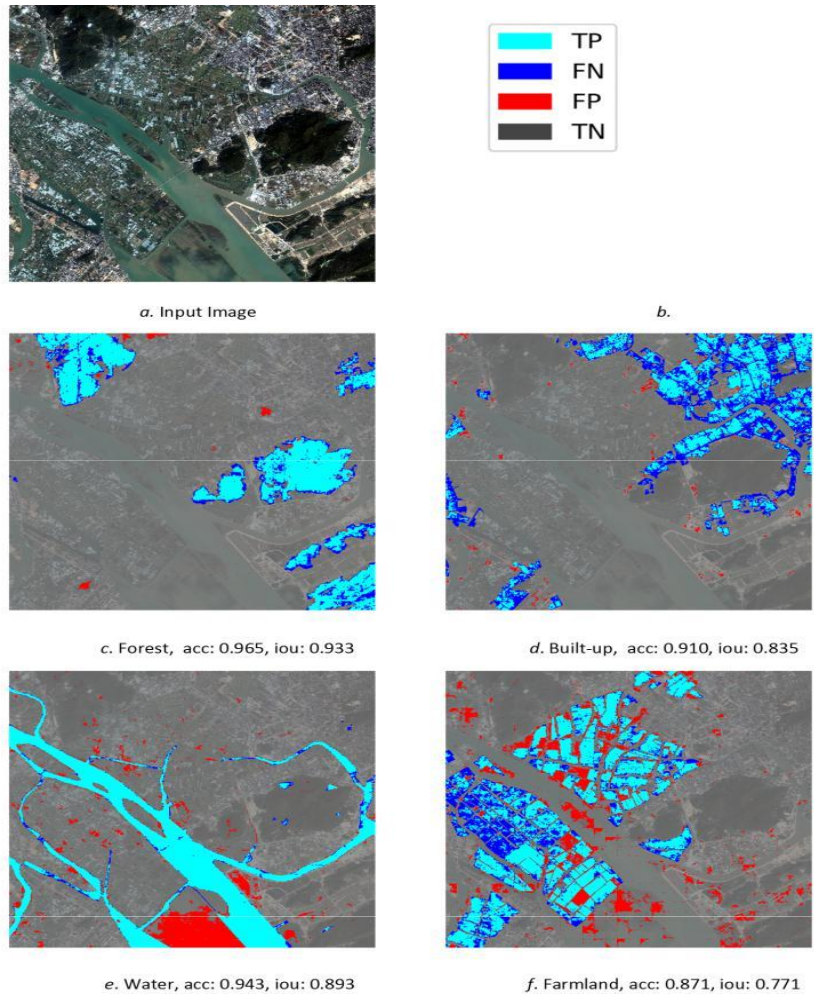

Fig. 5. Input image (a); (b) The legends. TP means true positive pixels, FN indicates false negative pixels, FP means false positive, and TN represent true negatives, (c) Forest output. cyan represents correctly classi ed forest area, blue pixels indicate false negative pixels. forest was there, but the model cannot detect forest, red means false positive pixels. forest was not there but the model predicts forest; Gray pixels act as the background pixels; (d) Built-up output, (e) Water output, (f) Farmland output

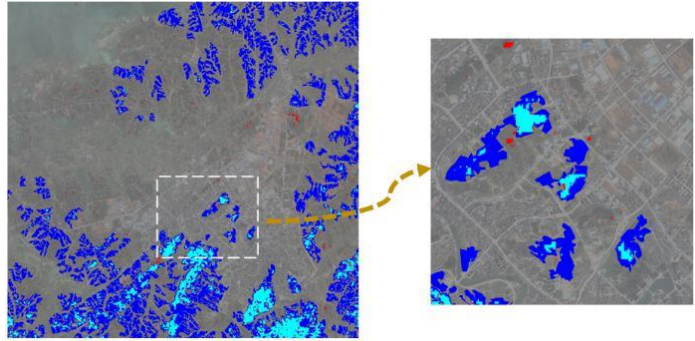

Fig. 6. (a) Output image for Forest segmentation; cyan represents correctly segmented forest area; blue indicates: forest was there, but the model cannot detect forest; red means: forest was not there but the model predicts forest; Gray means the background pixels; (b) zoomed in area of the dashed area of (a).

Accuracy and IoU for grid sub-image approach are signicantly improved. Among them, the accuracy scores for Forest, built-up, farmland, and water have been increased 26.07%, 34.79%, and 31.95%, respectively compared to eCognition.

### 4.4 Error Analysis

In this section, we try to explore where the CNN fails to segment and why. Figure 6a is the output of the FCN-8 model where cyan represents correctly segmented forest where we use downsampled image as input. While analyzing the performance, we are closely looking to the image 6a and the zoomed in version of in 6b where the segmentation fails severely. We made couple of observations to identify why pre-trained FCN-8 is failed to segment the forest accurately. There are very small regions with sharp boundaries present in the rectangle. Unfortunately, VGG-16 cannot model them unless the small region is big enough (6b blue region). We also observed that if a region is smaller than 10m ground sampling distance (GSD), the FCN-8 can not capture the region. We are zooming out the original image of 7168 6720 into 224x224 input images for the FCN-8. So, we are zooming out around 960 times smaller. As a result, the small regions that contain few pixels now lose the textures signicantly. Secondly, they lose their spatial GSD and LULC properties. Thus, this forest region (in an input image) already lose its ground region validity of a forest. As a result, the FCN-8 did not capture it as forest indeed. More importantly, the boundary pixels have less informative texture when we zoom out the images. Scaling down the original image massively reduces the discriminating features of the boundary pixels

These errors are alleviated by keeping the resolution of the images same while feeding them to the network. However, not all errors are xed by using the grid sub-images. We see that farmland has lower performance regarding the accuracy and IoU compared to other classes though we use grid sub-images. We found there are farmlands with very sharp edges in Figure 7a and 7b. Sharp boundaries are not captured by the FCN -8 model. Generally, FCN-8 always looks for smooth periphery. The FCN-8 is pretrained with ImageNet. And ImageNet does not have any objects that have such kinds of crisp boundaries. This might be the reason why the FCN-8 fails.

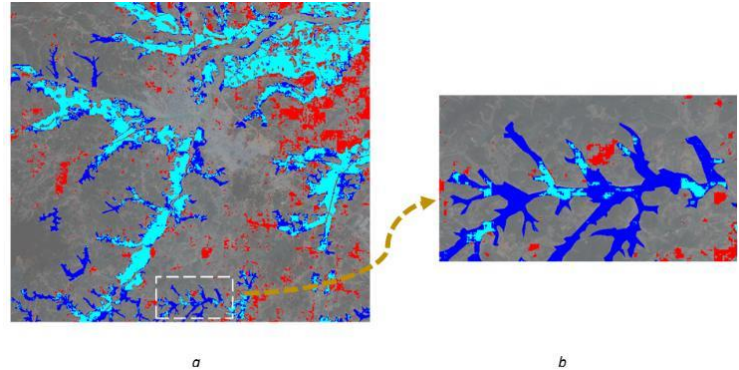

Fig. 7. (a) Output image for Farmland segmentation; cyan represents correctly segmented farm area, blue indicates: Farmland was there, but the model cannot detect farmland; red means: farmland was not there but the model predicts farmland; Gray acts as the background pixels; (b) zoomed in the dashed region.

## 5   Conclusion

In this paper, we propose a semantic segmentation framework using Fully Convo-lutional Network (FCN-8) to segment Land Use Land Cover classes from RGB satellite images only. We employed non-overlapping grid-based approach with FCN - 8 that obtains signi cantly improved performance than the GIS soft-ware. eCognition. The average accuracy is. 91.0% and the average Intersection over Union (IoU) is. 0.840. However, eCognition gets average accuracy. 74.0% and average IoU. 0.60 only. Our future work includes developing a new deep architecture for better boundary segmentation.

# Bibliography

[1] Badrinarayanan, V., Kendall, A., Cipolla, R.. Segnet. A deep convolutional encoder-decoder architecture for image segmentation. TPAMI 39(12)

[2] Benediktsson, J.A., Palmason, J.A., Sveinsson, J.R.. Classi cation of hyper-spectral data from urban areas based on extended morphological pro les. IEEE Transactions on Geoscience and Remote Sensing 43(3) (2005)

[3] Bioucas-Dias, J.M., Plaza, A., Dobigeon, N., Parente, M., Du, Q., Gader, P., Chanussot, J.. Hyperspectral unmixing overview. Geometrical, statistical, and sparse regression-based approaches. IEEE Journal of Selected Topics in Applied Earth Observations and Remote Sensing 5(2), 354{379 (2012)

[4] Cheng, G., Han, J., Guo, L., Liu, Z., Bu, S., Ren, J.. E ective and e - cient midlevel visual elements-oriented land-use classi cation using vhr re-mote sensing images. IEEE Transactions on Geoscience and Remote Sensing 53(8), 4238{4249 (2015)

[5] Hamida, A.B., Benoit, A., Lambert, P., Klein, L., Amar, C.B., Audebert, N., Lefevre, S.. Deep learning for semantic segmentation of remote sensing images with rich spectral content. In. 2017 IEEE International Geoscience and Remote Sensing Symposium (IGARSS). pp. 2569{2572. IEEE (2017)

[6] Li, J., Bioucas-Dias, J.M., Plaza, A.. Semisupervised hyperspectral im-age segmentation using multinomial logistic regression with active learning. IEEE Transactions on Geoscience and Remote Sensing 48(11) (2010)

[7] Liu, S., Qi, Z., Li, X., Yeh, A.G.O.. Integration of convolutional neural networks and object-based post-classi cation re nement for land use and land cover mapping with optical and sar data. Remote Sensing 11(6)

[8] Long, J., Shelhamer, E., Darrell, T.. Fully convolutional networks for se-mantic segmentation. In. 2015 IEEE Conference on Computer Vision and Pattern Recognition (CVPR). pp. 3431{3440 (2015)

[9] MRLC. Multi-resolution land cover characteristics consortium. https.//www.mrlc.gov/data (May 2019)

[10] Piramanayagam, S., Saber, E., Schwartzkopf, W., Koehler, F.. Supervised classi cation of multisensor remotely sensed images using a deep learning framework. Remote Sensing 10(9), 1429 (2018)

[11] Simonyan, K., Zisserman, A.. Very deep convolutional networks for large-scale image recognition. arXiv preprint arXiv.1409.1556 (2014)

[12] Tong, X.Y., Xia, G.S., Lu, Q., Shen, H., Li, S., You, S., Zhang, L.. Learning transferable deep models for land-use classification with high-resolution remote sensing images (2018)

[13] Trimble. ecognition developer edition. http.//www.ecognition.com/

[14] Volpi, M., Tuia, D.. Dense semantic labeling of subdecimeter resolution im-ages with convolutional neural networks. IEEE Transactions on Geoscience and Remote Sensing 55(2), 881{893 (2016)